\documentclass[journal,twoside, web]{IEEEtran}

\usepackage{amsmath,amssymb,amsfonts}
\usepackage{algorithmic}
\usepackage{graphicx}
\usepackage{algorithm,algorithmic}
\usepackage{hyperref}
 \usepackage{booktabs}
\usepackage{balance}
\usepackage{textcomp}
\begin{document}
\title{Exploring Bias and Prediction Metrics to Characterise the Fairness of Machine Learning for Equity-Centered Public Health Decision-Making: A Narrative Review}

\author{
    Shaina Raza\IEEEauthorrefmark{1}, 
    Arash Shaban-Nejad\IEEEauthorrefmark{2}, 
    Elham Dolatabadi\IEEEauthorrefmark{3}, 
    Hiroshi Mamiya\IEEEauthorrefmark{4}%
    \thanks{
        \IEEEauthorrefmark{1}Shaina Raza is with the Vector Institute for Artificial Intelligence, Toronto, ON, Canada (Corresponding author, email: shaina.raza@torontomu.ca).
    }%
    \thanks{
        \IEEEauthorrefmark{2}Arash Shaban-Nejad is with the Department of Pediatrics, University of Tennessee, Tennessee, United States (e-mail: ashabann@uthsc.edu).
    }%
    \thanks{
        \IEEEauthorrefmark{3}Elham Dolatabadi is with the School of Health Policy \& Management, York University and Vector Institute, Toronto, ON, Canada (email: edolatab@yorku.ca).
    }%
    \thanks{
        \IEEEauthorrefmark{4}Hiroshi Mamiya is with the Department of Epidemiology, Biostatistics, and Occupational Health, McGill University, Montréal, Québec, Canada (email: hiroshi.mamiya@mail.mcgill.ca).
    }
}

\maketitle

\begin{abstract}
\textbf{Background}: The rapid advancement of Machine Learning (ML) represents novel opportunities to enhance public health research, surveillance, and decision-making. However, there is a lack of comprehensive understanding of algorithmic bias — systematic errors in predicted population health outcomes — resulting from the public health application of ML. The objective of this narrative review is to explore the types of bias generated by ML and quantitative metrics to assess these biases.    
 
\textbf{Methods} : We performed search on PubMed, MEDLINE, IEEE (Institute of Electrical and Electronics Engineers), ACM (Association for Computing Machinery) Digital Library, Science Direct, and Springer Nature. We used keywords to identify studies describing types of bias and metrics to measure these in the domain of ML and public and population health published in English between 2008 and 2023, inclusive. 

\textbf{Results}:  A total of 72 articles met the inclusion criteria. Our review identified the commonly described types of bias and  quantitative metrics to assess these biases from an equity perspective.   

\textbf{Conclusion} : The review will help formalize the evaluation framework for ML on public health from an equity perspective.   
\end{abstract}


\section{Introduction}
\label{sec:introduction}
\subsection{Public Health Utilities of Machine Learning}
The aim of public health is to promote health and prevent disease, illness, and injury. 
The essential public health functions include health promotion, health surveillance, health protection, population health assessment, disease and injury prevention, and emergency preparedness and response \cite{CPHO2021}. Artificial Intelligence (AI) has demonstrated the potential to advance these key activities \cite{CPHO2021}.
Its key discipline, Machine Learning (ML) led to the emergence of algorithms that incorporate ubiquitous social, environmental and population health data and generate accurate predictions for community health status, such as disease outcomes and risk factors \cite{panch2019artificial, mhasawade2021machine, mooney2018big}. 

By integrating multiple data streams, ML algorithms can provide a comprehensive understanding of public health dynamics in real time \cite{raza2022coquad}. 
Public health applications of ML include computer vision techniques that identify urban environmental drivers of health, such as residential greenspace from a large number of street images \cite{larkin2021predicting}. ML has also been shown to be effective in identifying early warning signs of disease outbreaks such as COVID-19 \cite{shah2022prediction, raza2023connecting}, forecasting disease prevalence by incorporating local environmental data \cite{myers2000forecasting} or social media data \cite{dolatabadi2023}, and detecting outbreaks \cite{asif2020role},\cite{gao2020machine,cho2023detection}. Moreover, ML enables the large-scale and automatic assessment of behavioural risk factors, such as physical activities and dietary patterns by processing the stream of data generated by wearable devices \cite{teixeira2021wearable}. It also facilitates the population-scale understanding of public sentiment on various health topics from online social media data  \cite{white2023exploring}.  Finally, ML can support the development of targeted interventions tailored to the needs of specific populations \cite{mhasawade2021machine} and help optimize the allocation of finite public health resources to those who can benefit most from interventions\cite{mhasawade2021machine, ordu2021novel}.  

\subsection{Ethical Implications of Utilizing Machine Learning}
However, the integration of ML algorithms into public health research and practice imposes ethical challenges, as it can generate disproportionately inaccurate predictions of diseases and risk factors among disadvantaged and marginalized population subgroups \cite{mhasawade2021machine}. Such biased or ``unfair" prediction can lead to inappropriate public health interventions not adapted to the needs of disadvantaged populations. 

Health disparity represents disproportionately higher disease occurrences and risk factors, and lower resources to achieve health among disadvantaged populations \cite{carter2002health}. These vulnerable and equity-deserving populations are characterized by the social determinants of health, which include race, ethnicity, income, education, disability, access to healthcare, sex, and gender \cite{carter2002health, raza2023discovering}. These key population characteristics are often called  ``sensitive attributes" in the field of AI fairness. Equitable public health strategies are policies and programs that reduce these gaps by targeting equity-deserving groups, including racialized minorities, low-income and low-education individuals, and females and women. Thus, equity represents fair opportunities for all individuals to achieve optimum health by reducing existing health disparities.   


Bias poses a significant challenge to applying ML equitably in public health, given the  ML algorithm's  tendency to generate inaccurate prediction for health outcomes among certain population groups  \cite{panch2019artificial,raza2023fairness}. We note that the term ``bias'' has a slightly different definition in epidemiology, defined as a systematic deviation of etiologic associations from the (unobserved) true association \cite{ kleinbaum1981selection}. Such bias in epidemiological investigation arises from erroneous practice in data collection and participant recruitment/retention (i.e., selection bias), measurement error (i.e., information bias) and inappropriate data analysis (e.g., collider stratification bias) \cite{arnold2016brief,greenland1989ecological, halpern2020cognitive, johnson2008bias}. Throughout this study, we will use the former definition of bias related ML algorithms rather than epidemiological investigation, thereby focusing on the bias in prediction of population health (as opposed to bias in etiologic association).

\begin{figure}[h]
    \centering
    \includegraphics[width=1\linewidth]{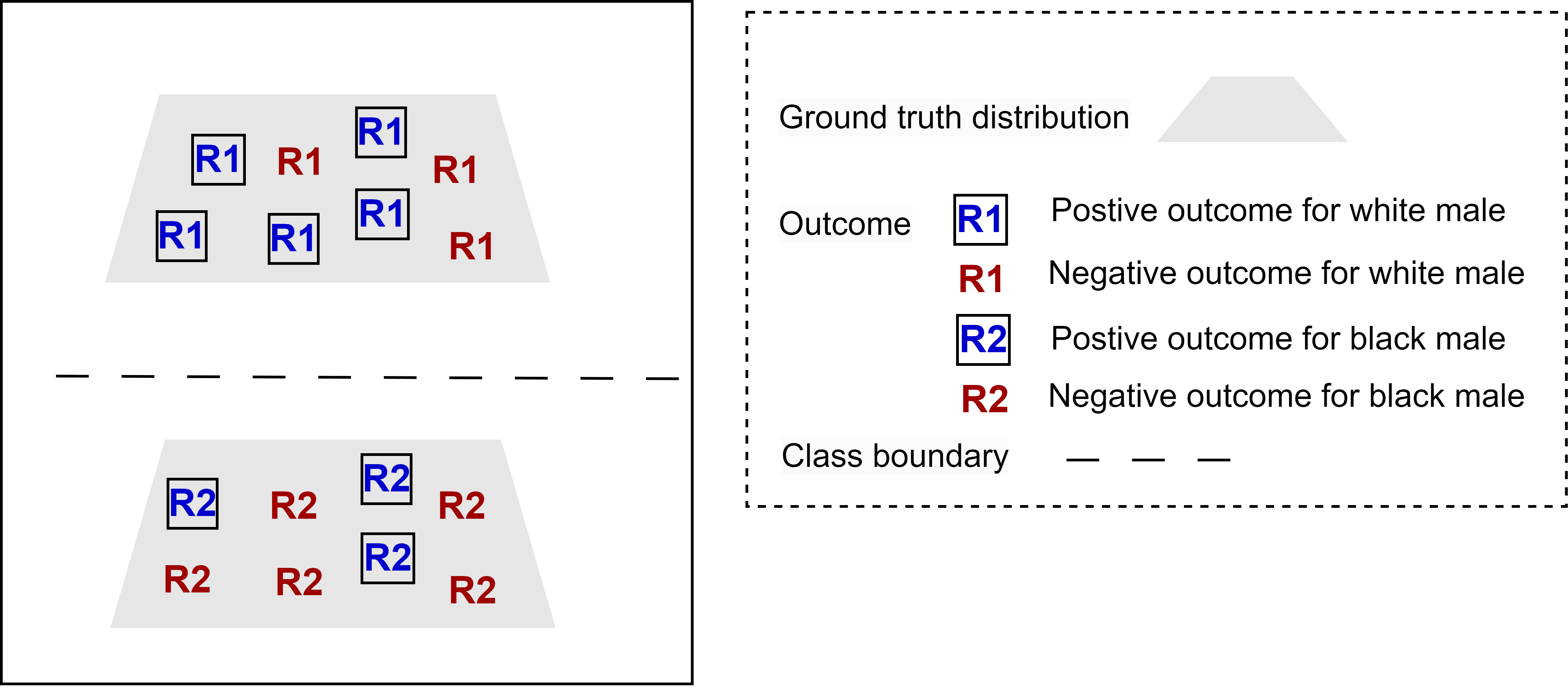}
\caption{Schematic Representation of Outcome Distribution by Race and Gender. This diagram illustrates the distribution of positive (blue and bordered) and negative (red) outcomes from an ML model for two groups, R1 and R2, representing white and black males, respectively. Notably, the classifier exhibits higher diagnostic accuracy for the condition in patients from the blue group R1 (White males) compared to the red group R2 (Black males). The dashed line represents the class boundary separating the outcomes for the two groups, visually emphasizing the disparity in accuracy between the groups.}
    \label{fig:fig1}
\end{figure}

To illustrate unfair predictions across different population subgroups due to bias, we present a scenario depicted in Figure \ref{fig:fig1}. This figure shows a binary classifier (a type of ML method to predict a binary health outcome e.g., disease diagnosis) designed to assess a specific health condition at the population level, utilizing patient data that includes social determinants of health, race in this example. The algorithm's accuracy is noticeably higher for the category of white male compared to others. This discrepancy underscores the classifier’s inability to equitably predict health outcomes, resulting in biases that favor certain groups at the expense of others. Such biases can often lead to serious public health issues, including delayed diagnoses and inappropriate treatments, which ultimately result in the widening of the existing disparities \cite{peterson2021health}.

\subsection{Fairness and Equity in Machine Learning Applications}
To mitigate bias, the data collection, development, evaluation, and implementation of ML should be conducted with fairness in consideration. This implies that assessing the potential impact of ML algorithms across populations at all stages of the ML lifecycle. Algorithmic fairness in AI is defined as the principle that decisions made by AI systems should not create unjust or prejudicial outcomes for certain groups based on their race, gender, or other social determinants of health (i.e., sensitive characteristics) \cite{verma2018fairness}. 

Research has demonstrated that a lack of fairness in AI decision-making processes can exacerbate inequities in public health outcomes. Extant studies \cite{mhasawade2021machine,park2022algorithmic,thomasian2021advancing} highlight how biases in AI systems contribute to unequal healthcare experiences and results among different population groups. In public health, the term ``fairness'' implies a condition that ensures impartial treatment of communities and individuals and is related to equity  \cite{galea2023within, braveman2006health}. While the definition of fairness is interrelated between AI and public health, our study largely focuses on the former definition (algorithmic fairness in AI), which centers on measuring the equitable impact of ML, with the goal of preventing lower accuracies of ML algorithms among equity-deserving population subgroups \cite{mitchell2021algorithmic}. 


\paragraph{Research Aim and Objective}
This study aims to review and summarize the existing types of algorithmic bias that characterize unfair prediction of ML from the literature. We also explore and introduce the existing metrics to measure the bias created by of ML.  The review adopts a narrative approach, which is suitable for broad topics.

Our study is unique to the existing reviews in AI fairness and medical and population science. The reviews are grouped into two broad categories. The first category of articles \cite{cordeiro2021digital,gervasi2022potential,rajkomar2018ensuring,sikstrom2022conceptualising,ahmad2020fairness,xu2022algorithmic,mccradden2020ethical} mainly discusses the potential benefits and risks of using ML and AI in healthcare, with largely theoretical (mathematical) discussion to derive algorithmic fairness . The second group of literature \cite{azimi2023optimizing,kulikowski2022ethics,lin2022preeclampsia,lu2022considerations,wesson2022risks,yang2022algorithmic,liu2024faircompass}  describes biases in healthcare data and ML algorithms, with general recommendations for incorporating an equity lens into big data research, data science, and performing fairness audits, rather than focusing on specific types of bias and fairness metrics.  Some other works \cite{dukhanin2018integrating,allin2005wanless,Marmot2010FairSH} discuss the challenges and opportunities in public health decision-making for policy objectives. 
Different from previous work, our review focuses on exploring the existing performance metrics of ML and biases that are directly relevant to measuring equity in population health. 

\section{Methods}

\begin{figure}[h]
    \centering
    \includegraphics[width=1\linewidth]{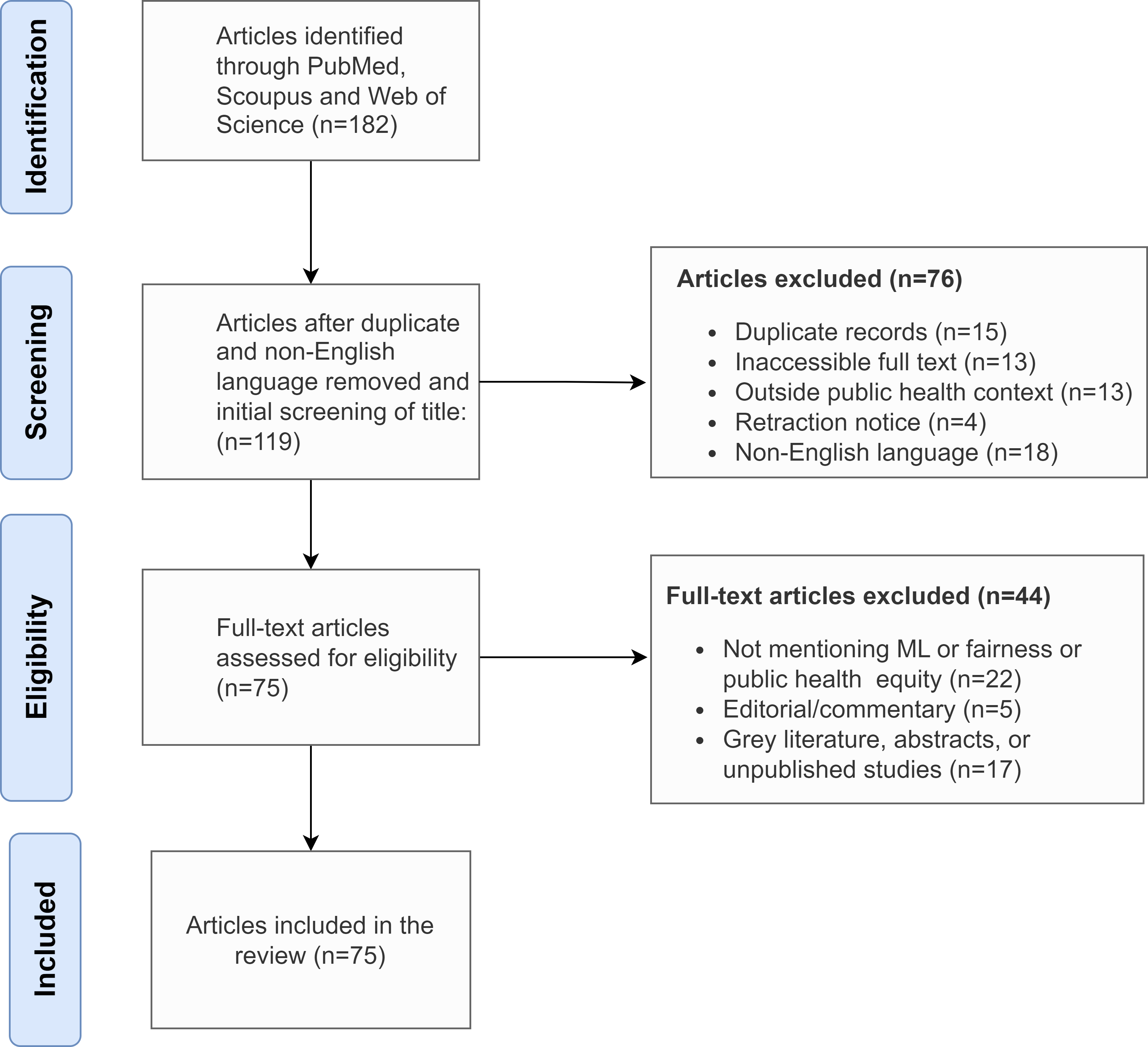}
    \caption{Flow diagram showing the process of study selection.}
    \label{fig:flowchart}
\end{figure}

In this work, we aim to address the question: "\textit{Which types of biases are prevalent in public health-related ML algorithms, and what quantitative fairness metrics can be used to identify and quantify these biases}?"
Our search consists of literature published from 2008 to 2023 to account for the limited volume of work at the intersection of these three fields: Public health, ML and equity.  The flow diagram showing the process of study selection is shown in Figure \ref{fig:flowchart}.
\subsection{Databases Used} 
We conducted a search across various electronic databases and digital libraries, including PubMed, MEDLINE, IEEE (Institute of Electrical and Electronics Engineers), ACM (Association for Computing Machinery) Digital Library, Science Direct, Springer Nature and Nature Portfolio, to identify studies focused on the intersection of public health equity and ML fairness, published in English and relevant to our research. Our approach spanned multiple disciplines to ensure a thorough review of quantitative fairness metrics in health-related ML algorithms.  


\subsection{Search Terms} 
We employed Medical Subject Headings (MeSH) and specific keywords such as: \\ `machine learning,' `public health,' and `equity,' using Boolean operators for precise refinement of search outcomes. The search syntax was structured as follows:   
(``Machine Learning'' [MeSH] OR ``machine learning'' OR ``deep learning'' OR ``neural networks'' OR ``artificial intelligence'') AND (``Public Health''[MeSH] OR ``public health'' OR ``health equity'' OR ``health disparities'' OR ``community health'' OR ``population health'' ) AND (``Bias''[MeSH] OR ``algorithmic bias'' OR ``fairness'' OR ``equity'' OR ``social determinants of health'' OR ``healthcare inequality'' OR ``predictive fairness'' OR ``ethical algorithms'') AND (``English''[Language]).

The inclusion criteria prioritized peer-reviewed articles that assessed or discussed quantitative fairness metrics in ML algorithms within the context of public health. Exclusion criteria were applied to non-peer-reviewed literature, non-English articles, and studies not focused on fairness metrics or their applications in public health.   

\subsection{Article Selection Strategy} 
The selection involved a preliminary review of titles and abstracts, followed by a full-text assessment for eligibility. Three independent reviewers conducted this process, resolving discrepancies by consensus or a fourth reviewer's input. A standardized form ensured consistent data collection, piloted on a subset of studies for reliability. 
We appraised study quality using a checklist adapted from the Joanna Briggs Institute (JBI) \cite{JBI2017} standards, assessing bias risk and methodological rigour to ascertain each study's contribution to our research question. This critical appraisal was independently executed by two reviewers, with a third resolving any differences.


\section{Results}
\subsection{Biases impacting fairness of ML algorithms}
\begin{figure}
    \centering
    \includegraphics[width=1\linewidth]{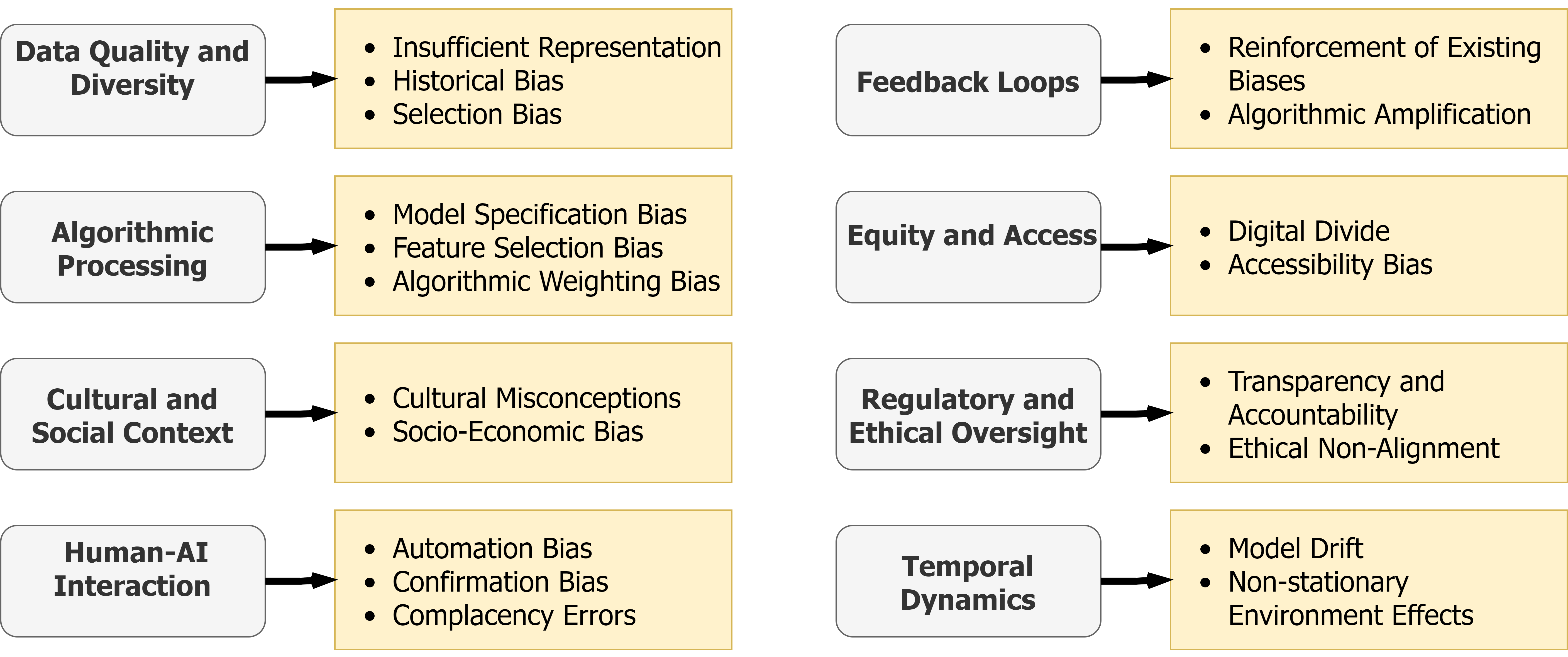}
    \caption{Various sources and categories of biases within the public health context mapped to concepts (rounded rectangles) }
    \label{fig:ph}
\end{figure}

We identified types of bias that are known to lead to unfair predictions by ML algorithms and grouped them into  potential sources of these biases, as listed in Figure \ref{fig:ph}  and are detailed below.

\subsubsection{Data Quality and Diversity} 
Data quality refers to minimizing errors in measurement, such as inaccuracies in labelling and data entry. This process is highly important to ensure that the data used in ML models accurately represent health trends and outcomes \cite{chen2014review}.  Data diversity refers to include the samples from all population groups.
Relevant research, such as  \cite{de2021developing,hardt2016equality,straw2020automation,kordzadeh2022algorithmic}, highlight this challenge of data quality and diversity. 
\\ 

\textit{Insufficient data representation} refers to the bias resulting from training datasets that do not adequately capture the diversity of the population. 
This lack of diversity in data can cause ML models to generate inaccurate predictions or decisions for underrepresented groups, exacerbating existing inequalities and compromising the ability to generalize across the entire population \cite{de2021developing}. For example, a ML model developed for medical diagnosis might be trained mostly on data from one ethnic group. This could result in less accurate diagnoses for patients from other ethnic backgrounds. 

\textit{Historical Bias }arises when ML models are trained on historical data embedded with systematic inequalities and societal biases. Such biases can lead to unequal treatment of specific groups (based on race, gender, age and so on) \cite{straw2020automation,kordzadeh2022algorithmic}. An example of historical bias can be found in health insurance systems. If the data used to train these systems includes biases against certain racial or ethnic groups, likely due to unequal access to healthcare or different treatment based on socio-economic status; the resulting ML models might continue to perpetuate these biases. As a result, people from these groups could be unfairly labeled as higher risk, leading to them facing higher insurance premiums or even being denied coverage, regardless of their actual health or needs.

\textit{Selection Bias} arises when the dataset used to train ML models does not accurately represent the broader population or specific scenario it aims to model. This form of bias often emerges from non-random sampling methods that inadvertently leave out significant portions of the population, resulting in ML systems that may not function equitably or effectively across various groups or conditions \cite{faugier1997sampling, tripepi2010selection, cortes2008sample}. An example of selection bias can be observed in facial recognition technology, where a ML model trained predominantly with images of individuals from certain racial or ethnic groups will likely exhibit enhanced performance in recognizing individuals from those groups \cite{parikh2019addressing}.


\subsubsection{Algorithmic Processing} 
Bias in ML predictions can occur during algorithmic processing or training. This includes biases like model specification bias, feature selection bias, and algorithmic weighting bias. These types of bias introduce systematic errors into AI systems. They affect the importance or weight assigned to different features and impact the fairness of decisions \cite{kordzadeh2022algorithmic,obermeyer2019dissecting,mickenautsch2010systematic,pudjihartono2022review,akay2009support,luyckx2015birth,ntoutsi2020bias}. These biases may initially appear during data collection and preparation due to sampling bias, where data might not fully represent the target population \cite{cortes2008sample}. To address this, techniques like using diverse datasets, balanced feature selection, and unbiased data labeling are encouraged (details in Additional File 1).

\textit{Model Specification Bias} Incorrect assumptions in models refer to biases introduced due to erroneous assumptions or oversimplifications made during the design phase of ML models. These biases can result in systematic errors in data processing and decision-making, which affects the model fairness and effectiveness across various scenarios and populations \cite{kordzadeh2022algorithmic,obermeyer2019dissecting,mickenautsch2010systematic}. For example, a medical imaging ML model trained mainly on data from Caucasian patients may struggle to diagnose diseases in Asian, African, or other descent patients accurately. This happens because the model incorrectly assumes that the characteristics of its training data apply to everyone. As a result, it can lead to higher misdiagnosis rates for underrepresented groups.
 
\textit{Feature Selection Bias }occurs when the data features (variables) chosen to train the model do not adequately represent the problem space or when irrelevant features are included that do not contribute to the model training to make accurate predictions. This bias can skew the outcomes of AI applications, and could lead to models that are less effective or fair \cite{pudjihartono2022review,akay2009support}. For example, a health AI application predicting patient disease risks may face feature selection bias if it relies on broad features like age and gender while ignoring important factors like lifestyle, diet, or genetics. This can lead the model to inaccurately assess disease risks, and can results in underestimations or overestimations for certain patient groups.

\textit{Algorithmic Weighting Bias} occurs when an ML model assigns disproportionate importance or ``weight" to certain features or variables over others in its decision-making process. This uneven weighting can significantly impact the outcomes of algorithmic decisions, potentially leading to unfair or biased results \cite{pudjihartono2022review,luyckx2015birth,ntoutsi2020bias}. For example, a ML model in patient triage systems might overly prioritize certain symptoms or demographic factors like age or gender, while not fully considering a patient’s complete medical history or current symptoms. This can result in misclassifying the urgency of care, causing younger patients to be prioritized over older ones or some symptoms being undervalued, irrespective of the true medical urgency.
 
\subsubsection{Cultural and Social Context}
Biases such as cultural misconceptions and socio-economic biases highlight the challenges in designing AI systems that are culturally sensitive and equitable across different socio-economic groups~\cite{dwivedi2021artificial,pickett2015income,raza2023discovering,ahnquist2012social}. Language and communication biases further demonstrate the difficulty in creating ML models that accurately understand and process language nuances and dialects~\cite{gala2023utility,jakesch2023human}. 

\textit{Cultural Misconceptions} in health contexts, occur when AI systems fail to understand or incorporate the cultural nuances, values, beliefs, and practices of diverse populations. This lack of cultural sensitivity can lead to inappropriate, ineffective, or even harmful health recommendations and diagnoses \cite{dwivedi2021artificial,raza2023improving}. For example, an AI telehealth platform trained primarily on white population data may struggle to accurately interpret symptoms from diverse cultural backgrounds. This could result in incorrect treatment recommendations, as the AI might miss key information when patients describe their conditions using culturally specific terms or expressions. 

\textit{ Socio-economic factors} also influence health outcomes, access to care, and the effectiveness of medical treatments. AI technologies that fail to account for these variables might exacerbate existing disparities \cite{pickett2015income,raza2023discovering,ahnquist2012social}. For example, an AI system designed to optimize patient flow in a hospital might prioritize patients based on factors like promptness of insurance approval or the ability to schedule follow-up visits. 
 
\subsubsection{Human-AI Interaction }
Biases in human-AI interactions, such as automation bias, confirmation bias, and complacency errors, highlight the complex relationship between humans and AI systems. These biases highlights the need for continuous human oversight and critical assessment to ensure that decisions made with AI assistance are both accurate and fair \cite{straw2020automation,kerasidou2022before,modgil2021confirmation,oswald2004confirmation,grissinger2019understanding}.

\textit{Automation Bias }highlights the risk of over-reliance on AI systems, potentially leading to the undervaluation of human expertise and oversight. This bias can emerge in any field where AI is employed, including healthcare, where it may influence decision-making processes or the interpretation of data \cite{straw2020automation,kerasidou2022before}. For example, if a ML model is highly accurate but still has a small margin of error, healthcare professionals might neglect to perform a thorough review or consider other diagnostic possibilities, relying only on the AI conclusion.
 
\textit{Confirmation Bias} is the tendency of humans to search for, interpret, favour, and recall information in a way that confirms or supports one's prior belief system or values \cite{modgil2021confirmation,oswald2004confirmation}. For example, users might frame queries to AI in a way that is likely to produce confirming evidence for their beliefs or may interpret AI outputs in a way that aligns with their preconceptions, as seen in interactions with large language models such as ChatGPT or Bard. 


\textit{Complacency errors} refer to the phenomenon where users become overly reliant on AI systems, leading to decreased vigilance and critical assessment \cite{grissinger2019understanding,kerasidou2022before}. This issue arises when individuals trust AI outputs without sufficient skepticism, potentially overlooking errors, biases, or inappropriate conclusions generated by the AI.

\subsubsection{Feedback Loops}
Feedback loops, such as reinforcing existing biases and algorithmic amplification, highlight the risk of perpetuating and exacerbating biases through AI systems~\cite{devillers2021ai,chapman2013physicians,hao2019ai,yang2022algorithmic}. 

\textit{Reinforcement of Existing Biases} In a standard AI setup, the outcomes may be shaped by pre-existing biases, which then get integrated into the system during subsequent training phases, and are amplified in model predictions \cite{devillers2021ai,chapman2013physicians}. For instance, if a ML model is trained on healthcare records that inadequately represent specific demographic groups, it could yield less precise or biased assessments for those populations. When such biased results are reintroduced into the training process, they amplify the original bias, initiating a loop that sustains and intensifies these biases over time.
 
\textit{Algorithmic Amplification }refers to the phenomenon where small initial biases in AI systems are magnified over time through iterative learning processes. This occurs because the AI system learning algorithms tend to give more weight to data patterns they encounter more frequently \cite{hao2019ai,yang2022algorithmic}. For example, a skin cancer ML model trained mostly on images of lighter-skinned patients might initially underperform on darker-skinned individuals. Over time, this bias can become more severe as the system keeps learning mainly from the data where it performs best (i.e., on the images of lighter-skinned patients).
 
\subsubsection{Equity and Access}
Issues such as the digital divide and accessibility bias highlight disparities in health outcomes and access to AI technologies. These disparities emphasize the need for equitable distribution and accessibility of AI resources across different populations to ensure fair health outcomes \cite{saeed2021disparities,campbell2012ethnic,whittaker2019disability,panch2019artificial}.

\textit{Digital Divide} refers to the gap between individuals with access to modern information and communication technology (ICT) and those without.  This divide can be due to various factors, including but not limited to socio-economic status, geographical location, age, and education level \cite{saeed2021disparities,campbell2012ethnic}. For example, urban hospitals use AI for early diabetes detection and personalized care, while rural areas lag, highlighting how the digital divide worsens health disparities.

\textit{Accessibility Bias} in AI implies that these technologies might not be designed with the needs of individuals with disabilities in mind, which might lead to unequal access \cite{whittaker2019disability,panch2019artificial}. For example, a voice-activated health app that does not offer text-based commands could exclude users with speech impairments and deny them the benefits of AI in monitoring health or accessing medical information.
 
\subsubsection{Regulatory and Ethical Oversight} 
Regulatory and ethical oversight challenges involve ensuring that AI systems maintain transparency, accountability, and adherence to ethical standards throughout their deployment \cite{williams2022transparency,buolamwini2018gender,gerke2020ethical,mccradden2020ethical}.

\textit{Transparency and Accountability} refer to making the actions in AI decision-making open and understandable. The goal is to ensure that individuals and organizations can be trusted and held responsible for their results \cite{williams2022transparency,buolamwini2018gender}.
. This challenge is highlighted by the difficulties patients and healthcare professionals face in understanding AI-generated diagnoses or treatment recommendations. Such challanges obstruct the broad acceptance and effective use of AI within the public health sector.

\textit{Ethical Non-Alignment} occurs when AI decisions in healthcare diverge from ethical norms or standards, potentially leading to scenarios where an AI recommends treatment options that, while clinically valid, may disregard patient values or ethical considerations \cite{gerke2020ethical,mccradden2020ethical}. For example, an AI system might suggest a less expensive treatment option with a slightly lower success rate over a more costly, but potentially more effective treatment, not considering the patient's preference for the best possible outcome despite the cost.

\subsubsection{Temporal Dynamics}
Temporal dynamics refer to the changes and trends over time that influence the behavior and outcomes of a system or process \cite{raza2019news}.  Temporal dynamics, like model drift and non-stationary environment effects, highlight the need for AI models to adapt continuously to evolving health profiles and emerging threats \cite{vela2022temporal,khan2023drawbacks}.

\textit{Model Drift }refers to the phenomenon where AI models become outdated due to changes in population health profiles over time. These drifts are often seen in the evolving disease patterns or emerging health trends that requires ongoing updates to maintain accuracy and relevance in healthcare predictions and treatments \cite{vela2022temporal}. For example, a ML model that was initially trained to predict flu outbreaks based on historical health data. If a sudden change in flu virus strains or a new pandemic emerges, like COVID-19, the model's predictions may become less accurate over time because it's not adapted to these new conditions.
 
\textit{Non-stationary Environment Effects} occur when AI systems struggle to adjust to dynamic health environments and emerging health threats that are not represented in the existing data \cite{khan2023drawbacks}. These effects can lead to decreased effectiveness in predicting and managing health conditions, as the models fail to incorporate the latest trends, outbreaks, or changes in population health behavior.

\subsection{Fairness Metrics to Assess the Performance of Machine Learning}

Fairness metrics are quantitative measures designed to capture and evaluate the fairness or equity of a system or decision-making process \cite{nielsen2020practical}. We identified accuracy metrics that facilitate the evaluation of the equitable performance of ML models. Sensitivity, analogous to recall, measures the proportion of actual positives correctly identified by the model, which is crucial for ensuring that no cases are missed in disease surveillance \cite{monaghan2021foundational}. Specificity, akin to the true negative rate, highlights the model's ability to accurately identify those population groups without the disease \cite{monaghan2021foundational}. The Positive Predictive Value (PPV) mirrors precision, indicating the proportion of positive test results that are truly positive, essential for assessing the model's accuracy in predicting disease presence \cite{monaghan2021foundational}. The F1 Score, which balances the PPV (precision) and sensitivity (recall) by penalizing imbalances, is crucial for models that need to correctly identify cases while avoiding false alarms. Finally, the overall accuracy metric reflects the model’s effectiveness in both disease detection and minimizing false alarms \cite{woodruff2009use}


\paragraph*{Common Fairness Metrics for Assessing Machine Learning Model Fairness in Public Health Contexts} The key fairness metrics used to evaluate ML models in public health contexts are as follows:

\begin{itemize}
    \item \textit{Disparate Impact Ratio}:   Disparate Impact Ratio is a quantitative measure used to identify and assess the level of disparity \cite{feldman2015certifying,zafar2017fairness}. It is calculated as the ratio of the rate at which a particular outcome occurs for a protected group to the rate at which that outcome occurs for the dominant or comparison group. A ratio of 1 indicates perfect equality; values less than 1 indicate that the outcome is less favourable for the protected group, and values greater than 1 indicate more favourable outcomes for the protected group. 

\textit{Example:} If an ML model is used to recommend mental health services and it is found that the urban population is twice as likely to be recommended for these services compared to the rural population, this could indicate a bias in how services are offered or accessed. Suppose $20\%$ of the urban population and only $10\%$ of the rural population receive enhanced mental health services recommendations. The Disparate Impact Ratio in this case would be $\frac{10\%}{20\%} = 0.5$. This ratio, being significantly less than $1$, suggests a disparate impact against the rural population, highlighting a potential bias in the algorithm's decision-making process.

\item \textit{Predicted Positive Rate (PPR)}: Measures the proportion of individuals within a specific group predicted by an ML model to have a positive outcome. This metric is crucial in evaluating the performance of predictive models across different demographic groups \cite{trevethan2017sensitivity}. \textit{Example:} When deploying a predictive model to identify patients at risk for heart disease, observing a PPR disparity; such as a PPR of 40\% for one ethnicity compared to a PPR of 20\% for another; can indicate potential biases or a lack of model sensitivity to diverse genetic backgrounds. Such disparities in PPR might suggest that the model disproportionately predicts positive outcomes (e.g., risk of heart disease) for one ethnic group over another, potentially due to factors like training data imbalances or overlooked genetic variations. 


\item \textit{False Discovery Rate (FDR)}: Evaluates the proportion of false positives among all positive predictions made by an ML model, offering insight into the model precision in identifying true positive outcomes \cite{storey2011false}. \textit{Example:} If the ML model designed to screen for stroke risk in patients, predicts that 100 individuals are at high risk of experiencing a stroke, but later assessments reveal that 40 of these predictions are incorrect (i.e., these individuals are not actually at high risk), the FDR for the model would be 40\%. This high FDR suggests a significant level of inaccuracy in the model's positive predictions, indicating a need for improvement in its predictive capabilities or a reconsideration of the features and data used for training to better distinguish between true and false positives \cite{grote2020ethics}.

\item \textit{False Positive Rate (FPR)}: This metric quantifies the frequency with which some 
individuals are incorrectly classified as having an outcome predictive model \cite{storey2011false}. \textit{Example:} If a cancer screening program using an ML model is designed to detect early signs of skin cancer and the program screens 1,000 individuals and mistakenly identifies 100 as at high risk of skin cancer, when in reality only 20 of these individuals exhibit early signs of the disease, the FPR would be calculated based on the number of false positives (those incorrectly identified as at risk) out of the total number of actual negatives (those who are healthy). In this case, if among the 1,000 screened, 980 are healthy, and 100 of these are falsely identified as at risk, the FPR would be $\frac{100}{980} \approx 0.102$, or 10.2\%. A high FPR, such as this, indicates that many individuals are wrongly subjected to the anxiety and unnecessary follow-up tests associated with a potential cancer diagnosis. This scenario expects a ML model to reduce false alarms and ensure that the screening program accurately distinguishes between healthy individuals and those with genuine disease.

\item \textit{False Omission Rate (FOR)}: This metric assesses the proportion of false negatives within all negative predictions made by a diagnostic test or predictive model \cite{riskyr}. \textit{Example:} In the context of screening for tuberculosis (TB), if a health screening program tests 1,000 individuals and 900 are predicted not to have TB (negative predictions). However, among these 900 individuals, 30 actually do have TB but were incorrectly classified as not having it (false negatives). The FOR in this case is calculated as the number of false negatives divided by the total number of negative predictions, which would be $\frac{30}{900} = 0.0333$ or 3.33\%. A high FOR in such a scenario suggests that a substantial number of TB cases might go undiagnosed due to the test or model's inability to identify all positive cases among those tested accurately.

\item \textit{False Negative Rate (FNR)}: This metric captures the proportion of positive cases that a diagnostic test or predictive model fails to identify \cite{riskyr}. \textit{Example:} In the context of COVID-19 testing, where a testing center conducts 1,000 tests, and there are 100 true positive cases among these. If the test fails to identify 20 of these true positives, reporting them as negative, the FNR would be calculated as the number of false negatives divided by the total number of positive cases, which would be $\frac{20}{100} = 0.20$ or 20\%. A high FNR in such a critical situation indicates that many individuals infected with COVID-19 are mistakenly informed they do not have the virus. This misdiagnosis can have severe public health implications, as these individuals, believing they are virus-free, might not take necessary precautions to isolate themselves, thereby risking the continued spread of COVID-19 to others. 


\item \textit{Group Size Ratio (GSR)}: This metric ensures the proportionate representation of different demographic or characteristic groups within a dataset or study population \cite{MathWorksRisk}. \textit{Example:} In a clinical trial investigating the effects of new diabetes medication, if there are 1,000 participants are enrolled, but only 200 of these are women. The GSR for women compared to men in this study can be calculated by dividing the number of female participants by the number of male participants. If there are 800 male participants, the GSR would be $\frac{200}{800} = 0.25$. This low GSR indicates a significant underrepresentation of female participants, which could lead to outcomes that might not be fully representative or equally applicable across genders.

\item \textit{Equality of Opportunity}: This metric ensures that individuals from different demographic or socio-economic groups have equal chances of receiving favorable outcomes, irrespective of their background \cite{roemer2015equality}. \textit{Example:} When evaluating the allocation of elective surgeries in a hospital, Equality of Opportunity can be applied to assess whether patients with varying insurance statuses are considered equally for surgery schedules. A review of hospital data reveals that patients with private insurance are twice as likely to be approved for elective surgeries compared to those with government-provided health coverage, even when controlling for medical needs. This discrepancy violates Equality of Opportunity, as the decisions seem to be influenced by the patient's ability to pay rather than their medical necessities.  

\item \textit{Equalized Odds}: This fairness metric targets the achievement of comparable error rates (both false positives and false negatives) across different demographic or characteristic groups \cite{awasthi2020equalized}. \textit{Example:} In diagnosing lung diseases, the diagnostic model sensitivity (true positive rate) and specificity (true negative rate) must be consistent across different patient groups, such as smokers and non-smokers.  If a lung disease diagnostic model is tested on a diverse patient group, analysis shows that the model has a higher sensitivity in detecting lung disease in smokers but a higher false positive rate in non-smokers. This discrepancy indicates that the model does not satisfy the Equalized Odds criterion, as the error rates (specifically, the rates of false positives and false negatives) vary significantly between these two groups. 

\item \textit{Balanced Accuracy}: This metric is designed to equalize the importance of sensitivity (the ability to correctly identify true positives) and specificity (the ability to correctly identify true negatives) in models, especially when dealing with imbalanced datasets or when the cost of different types of errors is similar \cite{brodersen2010balanced}. \textit{Example:} If a ML model demonstrates high sensitivity in detecting mental health needs in younger individuals but lacks specificity in older populations—resulting in many false positives among the older adults —the overall accuracy might appear acceptable. However, this overlooks the model's disproportionate error rates across age groups. Employing the balanced accuracy would highlight this disparity by giving equal weight to the model performance in correctly identifying both those who do and do not require interventions across all demographic groups.

\item \textit{Predictive Parity}: This metric ensures that the positive predictive value (PPV)—the probability that subjects with a positive screening test truly have the condition—remains consistent across different demographic or characteristic groups \cite{dieterich2016compas}. \textit{Example:} If a healthcare model designed to predict the likelihood of hospital readmission within 30 days of discharge, disproportionately flags elderly patients as being at a high risk of readmission compared to younger patients, despite similar health statuses and hospitalization histories, it may indicate a lack of Predictive Parity. Such a bias could lead to unnecessary stress for elderly patients and their families, increased healthcare costs, and inappropriate allocation of resources. To avoid these biased healthcare decisions, it is important to adjust the model to ensure that the likelihood of correctly predicting readmission is equally accurate across age groups.

\item \textit{Calibration}: This metric assesses how well the predicted probabilities from a model correspond to the actual outcomes, ensuring accuracy and reliability across different groups \cite{pleiss2017fairness}. \textit{Example:} If a surveillance system is designed to predict the risk of disease outbreaks, proper calibration means that if the model estimates a 30\% risk of an influenza outbreak in several regions, then, over time, approximately 30\% of those regions should indeed experience outbreaks. An analysis of the system reveals that while the predicted and observed outbreak patterns closely align in urban areas, the predictions are consistently overestimated for rural regions. This discrepancy indicates a lack of calibration, as the model's predictions do not accurately reflect the actual outcomes across different geographical areas. 

\item \textit{Balanced Error Rate (BER)}: This metric aims to achieve an equilibrium between the rates of false positives (incorrectly predicting a positive outcome when it is negative) and false negatives (failing to predict a positive outcome when it exists), thereby reducing bias in predictive modeling \cite{10.1214/09-AOS734}. \textit{Example:} In using predictive models for screening genetic disorders, the BER must be kept low and consistent across different ethnic groups. For instance, if a model is used to screen for a specific genetic disorder, it is found that the false positive rate is significantly higher in one ethnic group compared to others. In comparison, the false negative rate is higher in another group. Such discrepancies indicate a lack of balance in error rates, leading to potential bias against certain groups. 

\item \textit{Background Negative Subgroup Positive (BNSP) AUC}: This metric is designed to evaluate a model's accuracy in identifying positive cases within a predominantly negative background \cite{borkan2019nuanced}. \textit{Example:}In diagnosing a rare genetic disorder that affects only a small fraction of the population, the BNSP AUC metric would measure how effectively the model can pinpoint these few positive cases amidst a vast majority of negative (non-affected) cases. Achieving a high BNSP AUC is crucial because it signifies the model's capability to accurately identify the rare positive cases without generating an excessive number of false positives.

\end{itemize}


We present a case study on diabetes readmission based on these metrics in Additional File 1.

\section{Discussion}
This review article outlined the existing biases to describe the types of unfairness in the prediction of population health outcomes using ML algorithms. We also explored the existing quantitative metrics to identify and measure the extent of unfairness across population subgroups. Our findings will facilitate the use of these metrics to develop and evaluate of ML, thus promoting the equitable application of ML in public health. Our study also highlights biases occur at all stages of the ML lifecycle, including data processing and collection, model development, evaluation, implementation, and post-implementation.  

There is a guideline for the standardized development and reporting of medical prediction algorithms called Transparent Reporting of a multivariable prediction model for Individual Prognosis or Diagnosis’ (TRIPOD) \cite{collins2015transparent}, with a framework to evaluate and build statistical learning algorithms adapted to a clinical setting. However, it has a limited applicability to the assessment of fairness concerning ML algorithms and do not describe the specific types and sources of biases we provided. While the TRIPOD guideline adapted to ML algorithms is under development currently, our work will supplement such guideline by providing various fairness metrics adapted to the aim of each study concerning the equitable development and evaluation of ML. 

\paragraph{Limitations} 
Although we performed an extensive literature search, we might not have fully captured all recent studies, given the rapid advancement in the field of AI fairness and ML. Also, we note that the metrics for prediction performance, including the fairness metrics covered in our review, assume the accurate measurement of "ground truth", to which the predicted population health status is compared. However, systematic inaccuracy in ground truth measures routinely occurs due to bias in assessors and devices, for instance, assessment of medical charts or varying quality of devices that capture medical images \cite{zong2022medfair} across institutions or populations. Therefore, research efforts to increase the validity of ground truth measures is critical.
 

\paragraph{Future Directions}
Future research directions necessitate a multifaceted approach. First, there is a pressing need to cover studies that consider both the qualitative, besides the quantitative fairness methods. This should include a broader spectrum of ethical considerations relevant to AI applications in public health. Moreover, exploring innovative methodologies to enhance the effectiveness and applicability of fairness metrics across diverse public health contexts is crucial for effectively addressing and mitigating biases. Additionally, conducting a detailed case study is essential. These studies should delve into the practical challenges and opportunities of leveraging AI to promote public health equity, particularly capturing the diversity of public health scenarios and addressing the complexities of biases in real-world data.  

While we provide a comprehensive list of fairness metrics, the use of particular metrics should depend on the specific public health applications of ML algorithms. These applications include public health surveillance, outbreak control, health promotion, and resource prioritization and optimization.  Ideally, recommendations and frameworks to guide the appropriate fairness metrics suited for specific public health activities should be developed.  Finally, the application of the fairness metrics in real-world practice may be limited by the unavailability of the key variables characterizing equity-deserving variables, i.e., income, race, and gender, in the population data. The use of proxy measures for such critical attributes (e.g., aggregated census neighbourhood-level income) are often used but are likely to reduce the fairness of algorithms \cite{obermeyer2019dissecting}; thus, increasing access to these attributes at disaggregated, i.e., person-level, is critical, which can be achieved by the participation of public health and healthcare authorities for data sharing and linkage. 

\section{Conclusion}
This review examined fairness metrics and biases associated with the application of ML, highlighting the importance of using quantitative measures to identify and address biases effectively. The review points out the challenges of data biases and methodological limitations, stressing the need for interdisciplinary collaboration and strong ethical guidelines in ML an public health applications. The review results show ML’s potential to positively transform public health while warning of the risks of adopting methods without careful ethical and societal consideration.


\balance
\bibliographystyle{plain}
\bibliography{references}

\appendix



\section{Supplementary Information}
\subsection{Glossary of Terms in Public Health, Equity, Fairness, Bias, and Machine Learning}
\label{Appendix:terms}

 \textbf{Health}:  A state of complete physical, mental, and social well-being and not merely the absence of disease or infirmity. It covers the overall condition of an individual or population, including the presence or absence of illnesses, the level of physical and mental function, and the ability to respond to various factors that affect health.

\textbf{Public Health}:  Refers to the science and art of preventing disease, prolonging life, and promoting health through organized efforts and informed choices of society, organizations, public and private, communities, and individuals.

\textbf{Equity: } The absence of avoidable or remediable differences among groups of people, whether those groups are defined socially, economically, demographically, or geographically. In health care, equity refers to providing care that does not vary in quality because of personal characteristics such as gender, ethnicity, geographic location, and socio-economic status.

\textbf{Machine Learning (ML)}:  A subset of artificial intelligence that involves the development of algorithms and statistical models that enable computers or programs to perform a task without using explicit instructions, relying instead on patterns and inference derived from data.

\textbf{ Fairness: } In the context of ML and decision-making systems, fairness refers to the attribute that an algorithmic decision or model does not create biased outcomes, ensuring that all individuals or groups are treated equally and justly based on the relevant criteria.

\textbf{Bias:}  Refers to a systematic error or non-random distortion in results or inferences from processes. In data science and machine learning, bias can emerge due to prejudiced assumptions during the model development phase, from the data collection process, or in the algorithmic processing, leading to unfair outcomes.

\textbf{Algorithmic Bias: } Occurs when a computer program or algorithm reflects the implicit values of the humans who are involved in coding, collecting, selecting, or using data to train the algorithm. This can lead to skewed outputs and discriminatory practices.

\textbf{Social Determinants of Health (SDOH)}:  The conditions in the environments where people are born, live, learn, work, play, worship, and age that affect a wide range of health, functioning, and quality-of-life outcomes and risks.

\textbf{Health Disparities}:  A type of difference in health that is closely linked with social, economic, and/or environmental disadvantage. Health disparities adversely affect groups of people who have systematically experienced greater obstacles to health based on their racial or ethnic group; religion; socio-economic status; gender; age; mental health; cognitive, sensory, or physical disability; sexual orientation or gender identity; geographic location; or other characteristics historically linked to discrimination or exclusion.

\textbf{Predictive Analytics: } The use of data, statistical algorithms, and machine learning techniques to identify the likelihood of future outcomes based on historical data. In public health, it's used for predicting disease outbreaks, health trends, and patient outcomes.

\textbf{Informed Consent}:  A process for getting permission before conducting a healthcare intervention on a person. In research, it involves informing participants about the risks, benefits, and potential outcomes of studies or data collection.

\textbf{Privacy:  }In the context of data and digital tools, privacy concerns the right of individuals to control or influence what information related to them may be collected and stored and by whom and to whom that information may be disclosed.

\textbf{Data Governance: } The overall management of the availability, usability, integrity, and security of the data employed in an organization, with the objective to ensure that data is accurate, available, and accessible.
    
\textbf{Disparate Impact: } Refers to practices or policies that may appear neutral but result in a disproportionate impact on a protected group. It's often used to identify unintentional biases in systems or models.

\textbf{Predicted Positive Rate (PPR)}:  Measures the proportion of individuals in a group predicted to have a positive outcome. It's useful for identifying bias in predictive models across different demographic groups.

\textbf{False Discovery Rate (FDR):}  The proportion of false positives within all positive predictions. A high FDR indicates many incorrect positive predictions, suggesting a need for model recalibration.

\textbf{False Positive Rate (FPR): } Indicates how frequently healthy individuals are mistakenly identified as diseased. A high FPR can lead to unnecessary stress and testing, highlighting the need for accurate diagnostic criteria.

\textbf{False Omission Rate (FOR)}:  Assesses the rate of false negatives among negative predictions. A high FOR, especially in diagnosing infectious diseases, could lead to untreated spread within communities.

\textbf{False Negative Rate (FNR): } Reflects the proportion of actual positives missed by the model. A high FNR, such as in COVID-19 testing, means many infected individuals are wrongly reassured they are virus-free, posing a public health risk.

\textbf{Group Size Ratio (GSR): } Examines the representation ratio of different groups within the data. It's important for ensuring research outcomes are representative and equally applicable across all groups.

\textbf{Equality of Opportunity:}  Ensures individuals from different groups have equal chances of receiving a favorable outcome. It's crucial in healthcare to ensure decisions are based on medical need rather than the ability to pay.

\textbf{Equalized Odds:}  Aims for similar error rates across groups to ensure fairness in treatment recommendations and other decisions made by predictive models.

\textbf{Balanced Accuracy:}  Balances sensitivity and specificity across groups to ensure that a model accurately identifies needs regardless of demographic factors.

\textbf{Predictive Parity:}  Measures consistency in predictive outcomes across different groups, ensuring that no group is disproportionately flagged by predictive models.

\textbf{Calibration: } Evaluates the alignment of predicted probabilities with actual outcomes across groups. It's essential for maintaining the reliability of predictive models in public health and other areas.

\textbf{Balanced Error Rate (BER):}  Seeks to balance false positives and negatives to prevent bias in predictive modeling, ensuring equitable outcomes across all groups.

\textbf{Background Negative Subgroup Positive (BNSP) AUC:}  Focuses on model accuracy when identifying positive cases against a largely negative backdrop. It's critical in rare disease identification to avoid overwhelming false alerts.

\subsection{Algorithmic Processing}
\label{Appendix:algo}
Techniques such as data augmentation and synthetic data generation can improve data diversity~\cite{leavy2018gender,zhao2018gender,yang2022algorithmic}. Additionally, methods like re-weighting and re-sampling are critical for achieving fair representation~\cite{zelaya2019towards,kamiran2012data,zhang2020fairness}.

The model training phase is also susceptible to bias through the selection of model architecture, hyperparameters, and loss functions. 

Mitigation strategies include designing models less sensitive to biased features, using fair loss functions, and adjusting training processes to improve fairness~\cite{nielsen2020practical,lahoti2020fairness,mandal2020ensuring,kearns2018preventing,agarwal2019fair}. During model calibration, ensuring accurate and fair probabilistic estimates is crucial to avoid biases impacting prediction quality and decision fairness. Approaches for addressing calibration biases involve adjusting output probabilities, equalizing decision thresholds across diverse groups, and properly calibrating model outputs~\cite{hardt2016equality,jang2022group,pleiss2017fairness}.\\
\subsection{Equitable Predictions in Machine Learning: A Case Study on Diabetes Readmission}
\label{Appendix:case-study}
\subsubsection{Task Setting and Evaluation for Fairness}
This case study investigates the fairness of machine learning (ML) model predictions across various demographics—age, gender, and race—using a comprehensive diabetes dataset. Our primary focus is on evaluating the equity of these predictions in determining the risk of 30-day hospital readmissions for diabetic patients. We examine whether ML models offer equitable predictions across different demographic groups, particularly in forecasting 30-day readmissions among diabetes patients. Our objective encompasses not only assessing the accuracy of these models but also scrutinizing their fairness and potential biases to ensure that healthcare predictive analytics equitably benefit all patient groups.

Leveraging data from 130 US hospitals between 1999 and 2008, this study narrows down on 45,715 patient records from an initial pool of 101,766, after filtering out identifiers, features with substantial missing values, and records lacking race or diagnosis information. The final dataset includes 50 attributes, highlighting a class imbalance in the 30-day readmission indicator. Our analysis deploys four ML models—Multi-Layer Perceptron (MLP), Generalized Linear Model (GLM), and Naive Bayes—optimized for accuracy and fairness across demographic lines through specific hyperparameter settings.

The evaluation framework combines accuracy metrics (precision, recall, F1 score, and overall accuracy) with a thorough fairness analysis. By adopting metrics like the disparate impact ratio and predicted positive rates, among others, we aim to provide equitable predictions across all demographics. The study follows a structured evaluation flow: generating model predictions, assessing accuracy, conducting a fairness analysis for sub-groups, examining accuracy within these groups, and iteratively refining models to balance accuracy with fairness.

In a city, over the past year there is a hospital where 10,000 White patients and 5,000 Black patients sought diagnosis for a disease that requires hospitalization. The White population is assumed to be in the majority. We want to examine the privileged and underprivileged groups who get the favourable outcome (hospitalization here), regardless of their population size. The following confusion matrices represent the actual number of patients from each racial subgroup who were diagnosed with the disease and required hospitalization (positive label for ‘hospitalized’ and negative for ‘not hospitalized’).

\subsubsection{Results}
\begin{table}[h]
\label{tab:acc}
\small
\centering
\caption{Comparison of Different Models: Multilayer Perceptron (MLP), Generalized Linear Model (GLM), Naive Bayes (NB) Performance Across Precision, Recall, F1 Score, and Accuracy Metrics.}
\begin{tabular}{lcccc}
\toprule
Model & Precision & Recall & F1 Score & Accuracy \\
 MLP & \textbf{0.7978}& \textbf{0.987}& \textbf{0.882}&\textbf{0.819}\\
GLM & 0.7929 & 0.928 & 0.855& 0.8097 \\
Naive Bayes & 0.7786 & 0.823 & 0.801& 0.7404 \\
\bottomrule
\end{tabular}
\end{table}

Table \ref{tab:acc} compares three models (MLP, GLM, Naive Bayes) on metrics like Precision, Recall, F1 Score, and Accuracy. We observe that MLP performs best across all metrics. It indicates that MLP is most effective in balancing correct predictions and minimizing false positives and negatives. GLM follows closely, while Naive Bayes lags behind, possibly due to its assumption of feature independence. The choice of model should also consider factors like computational cost and interpretability beyond these metrics.

While the performance metrics of these models are key to measure their effectiveness, assessing their impact across various demographic groups is vital for ensuring fairness. Fairness entails that these models' predictions do not systematically advantage or disadvantage any group based on sensitive attributes like age, gender, race, or socio-economic status. Evaluating model performance with a lens on demographic equity is essential to uncovering biases that could result in unfair outcomes, thus ensuring the models operate equitably without reinforcing or worsening existing disparities.

\begin{table}[h]
\label{tab:fairness}
\centering
\small
\caption{Evaluation of Fairness Metrics (Predicted Positive Group Rate (PPGR), False Positive Rate (FPR), False Omission Rate (FOR), False Negative Rate (FNR)) Across Gender (Female and Male) Groups for MLP, GLM, and NB Models. Bold indicates the best score within each gender group, with lower error rates (FPR, FOR, FNR) preferred and higher PPGR values suggesting more equitable positive predictions.}
\begin{tabular}{lccccc}
\toprule
Gender & Model & PPGR & FPR & FOR & FNR \\
\midrule
Female & MLP & \textbf{0.37} & 0.16& 0.25 & 0.31 \\
       & GLM & 0.38 & 0.17 & 0.27 & 0.32 \\
       & NB  & 0.35 & 0.13 & 0.35& 0.45\\
\hline
Male   & MLP & \textbf{0.38} & \textbf{0.07} & \textbf{0.23}& \textbf{0.27}\\
       & GLM & 0.42 & 0.08 & 0.24 & 0.29 \\
       & NB  & 0.4  & 0.12 & 0.31& 0.42\\
\bottomrule
\end{tabular}
\end{table}

Table \ref{tab:fairness} presents a nuanced view of fairness metrics across gender groups for different models, revealing disparities in model biases. For females, the MLP model demonstrates a balance in predicted PPGR and maintains lower error rates compared to GLM and NB, indicating a slightly less biased performance towards female subjects. However, for males, the MLP model shows superior fairness metrics, with the highest PPGR and the lowest rates of FPR, FOR, and FNR, suggesting it is the least biased model across both genders. Conversely, the NB model exhibits higher error rates for both genders, particularly in the FOR and FNR for females, and overall for males, pointing to a greater bias. This analysis indicates that while no model is entirely free of bias, MLP appears to be the most equitable, whereas NB could be considered the most biased, with GLM occupying a middle ground.
\begin{table}[h]
\small
\centering

\caption{Comparative Analysis of Fairness Metrics by Race for MLP, GLM, and NB Models. The table highlights disparities in model performance across racial groups, focusing on fairness through PPGR, FPR, FNR and FOR metrics. Bold figures indicate the most favorable outcomes, with lower error rates (FPR, FOR, FNR) preferred and higher PPGR values suggesting more equitable positive predictions}
\begin{tabular}{lccccc}
\toprule
Race & Model & PPGR & FPR & FOR & FNR \\
\midrule
African American & MLP& 0.38 & 0.08 & 0.24 & 0.29 \\
& GLM & 0.39 & 0.08 & 0.24 & 0.29 \\
& NB & 0.36 & 0.14 & 0.31 & 0.40\\
\hline
Asian & MLP& \textbf{0.56}& \textbf{0.01}& \textbf{0.23}& \textbf{0.17}\\
& GLM & 0.41 & 0.05 & 0.27 & 0.29 \\
& NB & 0.36 & 0.04 & 0.48 & 0.46 \\
\hline
Caucasian & MLP& 0.37 & 0.06 & 0.24 & 0.31\\
& GLM & 0.39 & 0.07 & 0.24 & 0.29 \\
& NB & 0.36 & 0.13 & 0.31 & 0.41 \\
\hline
Hispanic & MLP& 0.38 & 0.11 & 0.35 & 0.39 \\
& GLM & 0.39 & 0.08 & 0.31 & 0.35 \\
& NB & 0.34 & 0.11 & 0.39 & 0.47 \\
\hline
Other & MLP& 0.5 & 0.11 & 0.34 & 0.27 \\
& GLM & 0.41 & 0.13 & 0.30& 0.34 \\
& NB & 0.45 & 0.11 & 0.40& 0.35 \\
\bottomrule
\end{tabular}
\label{tab:race}
\end{table}

Table \ref{tab:race} disparities in model performance across racial groups, focusing on fairness metrics. Asians benefit most notably from the MLP model's precision, experiencing the most fir treatment with the lowest FPR and highest PPGR, highlighting its superior predictive equity. In contrast, the Hispanic and Asian groups face the greatest challenges with the NB model, which shows a tendency to underpredict positive outcomes for these demographics, indicating significant biases.

Overall, in these results, we observe that the MLP model emerges as the most equitable, especially for Asians, by achieving an optimal balance between fairness and accuracy across racial groups. The GLM model displays consistent fairness metrics, offering a stable, albeit less exceptional, performance that does not heavily favor or disadvantage any particular group. On the other hand, the NB model requires refinement to address its bias, particularly in improving fairness for Hispanic and Asian populations, where it currently lags in predictive accuracy and equity.\label{Appendix:cast-study}

\end{document}